\documentclass[11pt,a4paper,table,xcdraw]{article}
\usepackage[draft]{hyperref}
\usepackage[hyperref]{emnlp2018}
\usepackage{times}
\usepackage{latexsym}
\usepackage{mathtools}
\usepackage{graphicx}
\usepackage{dblfloatfix}    
\usepackage{algpseudocode}
\usepackage{algorithm}
\graphicspath{ {images/} }
\usepackage{xcolor}
\usepackage{multirow}
\usepackage{booktabs}
\usepackage{url}
\usepackage{scrextend}
\usepackage{amsmath,amssymb}
\aclfinalcopy 

\title{Extractive Adversarial Networks: High-Recall Explanations for Identifying Personal Attacks in Social Media Posts}

\author{Samuel Carton \\
  University of Michigan \\
  {\tt scarton@umich.edu} \\\And
  Qiaozhu Mei\\
  University of Michigan \\
  {\tt qmei@umich.edu} \\\And
    Paul Resnick \\
  University of Michigan \\
  {\tt presnick@umich.edu} \\}

\date{}

\begin{document}
\maketitle

\begin{abstract}
  
We introduce an adversarial method for producing high-recall explanations of neural text classifier decisions. Building on an existing architecture for extractive explanations via hard attention, we add an adversarial layer which scans the residual of the attention for remaining predictive signal. Motivated by the important domain of detecting personal attacks in social media comments, we additionally demonstrate the importance of manually setting a semantically appropriate ``default'' behavior for the model by explicitly manipulating its bias term. We develop a validation set of human-annotated personal attacks to evaluate the impact of these changes.
\end{abstract}


\begin{figure*}[t]
\centering
\includegraphics[width=6in]{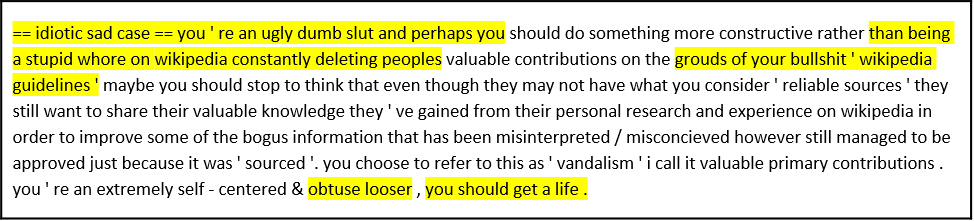}
\caption{An example of a highly-attacking comment from the test set, rationalized by the model}
\label{fig:high_attacking_example}
\end{figure*}

\begin{figure*}[b]
\centering
\includegraphics[width=6in]{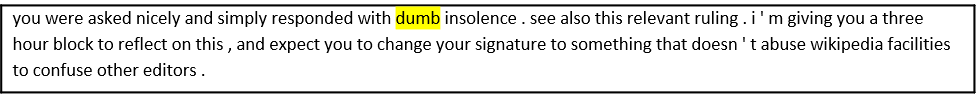}
\caption{An example of a not-very-attacking example from the test set, rationalized by the model}
\label{fig:low_attacking_example}
\end{figure*}

\section{Introduction}

The task of explaining classifier decisions has recently attracted increased attention from the research community. It is important for several reasons, including: 1) The increasing performance gap between simple-and-interpretable models and complex-but-opaque models (which demand more sophisticated explanation techniques); 2) The increasing ubiquity of machine learning in business and government and the concomitant need to understand the decisions of models in high-stakes situations; and 3) A rising awareness of the limitations of machine learning and the need for ways to better utilize intrinsically unreliable models (whose weaknesses can potentially be ameliorated by good explanations).

A common way to explain why a model classified an example a certain way is to extract a sparse subset of features that were responsible for the model's decision, sometimes described as a saliency mask or ``rationale'' in the case of text \citep{guidotti_survey_2018}. This type of local explanation may not completely elucidate why a given example is assigned a given outcome, but it does simplify the relationship by identifying what attributes were considered in the decision. 

Existing work on this topic has not explicitly addressed the problem of local feature redundancy. That is, when two features are equally predictive of an outcome, which of them should be included in the saliency mask for that decision? Typical sparsity constraints encourage minimal sufficient masks--unveiling just enough of the example to justify the outcome.

There are domains, however, where it may be important to produce complete explanations rather than minimal explanations. One example is the task of detecting content in online social media that violates a platform's policies. Explanatory models can potentially help human moderators make quicker and more consistent decisions about whether to remove comments \citep{lakkaraju_interpretable_2016}. However, we propose that truly minimal explanations are liable to give only a partial portrait of why a comment is objectionable, making it harder to render a fair holistic decision. If used to explain to a poster why their post was removed, a minimal explanation can actually be misleading, by implying that some of what was objectionable about their post was benign just because it didn't add marginal signal to the overall classification.

We use an extractive explanatory neural network to identify which social media comments contain personal attacks and which words in those comments are the basis for classifying them as containing personal attacks. We train this model on a large dataset \citep{wulczyn_ex_2017} of comments labeled for the presence of such attacks, and use the explanatory capacity of the model to identify spans that constitute personal attacks within those comments. We extend the work of \citep{lei_rationalizing_2016} in using one recurrent neural net (RNN) to produce an explanatory hard-attention rationale and a second RNN to make a prediction, the two models trained in an end-to-end fashion. 

To produce complete (i.e high-recall) explanations, we add to this existing architecture a second, adversarial predictive layer whose purpose is to try to make predictions based on what is left out of the rationale. We then add a term to the attention layer objective function which encourages it to fool this secondary predictive layer into making poor predictions by including all predictive signal (i.e personal attacks) in the mask that it generates.

We also show that manipulating the model bias term to set a semantically appropriate ``default behavior'' or ``null hypothesis'' for the model significantly improves performance. That is, by explicitly choosing what output a zero-information, empty explanation should correspond to, the model is able to learn explanations that correspond more closely with human-generated data. 

To summarize, the contributions of this paper are as follows:
\begin{itemize}
    \item We articulate explanation as an adversarial problem and introduce an adversarial scheme for extraction of complete (high-recall) explanations for text classifier decisions.
   \item We demonstrate the value of explicitly setting a default output value in such an explanatory model via bias term manipulation.  
   \item We apply explanatory machine learning for the first time to the task of detecting personal attacks in social media comments, and develop a validation dataset for this purpose. 

\end{itemize}

\section{Related work}
\subsection{Online abuse}
Online abuse (of which personal attacks are a major dimension) has recently attracted increased attention as a computational problem. Scholarly work has assessed the prevalence and impact of such abuse \citep{lenhart_online_2016,anderson_nasty_2014,pew_political_2016,anderson_toxic_2016}, while several initiatives have sought to construct datasets for its study \citep{wulczyn_ex_2017,abbott_internet_2016,kennedy_technology_2017,napoles_finding_2017,golbeck_large_2017}. 

Naturally, much recent work has gone into the use of machine learning to detect online abuse and its perpetrators \citep{nobata_abusive_2016,pavlopoulos_deep_2017,cheng_antisocial_2015}, including a workshop at the most recent ACL conference \citep{association_for_computational_linguistics_proceedings_2017}. However, the idea of fully-automated moderation by machine learning has attracted criticism as being subject to bias, inaccuracy, manipulation and frustration on the user end \citep{pavlopoulos_deep_2017,binns_like_2017,blackwell_classification_2018,hosseini_deceiving_2017,adams_better_2017}. We propose interpretable models as one potential solution to some of these problems.

\subsection{Interpretable machine learning}
Major points of division in the interpretability literature include: 1) local vs. global interpretability; 2) post-hoc vs. built-in interpretability; 3) explanation type; 4) input data type; and 5) evaluation metric. Our model is a built-in, feature-based locally interpretable model for text that we evaluate relative to a human gold-standard. \citet{guidotti_survey_2018} provides a recent survey of the field.

Recent work on interpretability has focused on local (i.e. instancewise) feature-based explanations. Attention models implicitly produce this type of explanation in the form of attention weights over input features. \citet{lei_rationalizing_2016} utilizes a regularized hard attention mechanism to identify the locally minimum sufficient subset of tokens to make accurate predictions. 

Post-hoc methods seek to retroactively probe the behavior of an existing non-explanatory model. These include model-specific gradient-based attribution methods, pioneered by \citet{simonyan_deep_2013} and reviewed recently by \citet{ancona_towards_2017}, which have tended to originate in the image classification domain and transfer to other domains such as text (e.g. \citet{arras_explaining_2017}). Conversely, LIME \citep{ribeiro_why_2016} is a prominent recent model-agnostic work in this space, building local linear approximations of a model and using the coefficients thereof to explain its behavior. \citet{li_understanding_2016} trains a hard attention layer to flip the decisions of an existing model. 

While feature-based explanations are the most common approach, other forms have been proposed, including: similarity to learned "prototypes" which represent clusters of items from the training data \citep{li_deep_2017-1}; high-precision feature interaction rules \citep{ribeiro_anchors:_2018}; reference to predefined human-friendly concepts \citep{kim_interpretability_2017}; and generated natural language \citep{ehsan_rationalization:_2017}. Likewise, many evaluation criteria have been proposed. These include fully automated evaluation \citep{arras_explaining_2017}; comparison to human gold standards \citep{lei_rationalizing_2016}; and human task performance \citep{nguyen_comparing_2018}. \citet{doshi-velez_towards_2017-1} and \citet{gilpin_explaining_2018} both present reviews and taxonomies of evaluation types.

\subsection{Adversarial learning}
Generative Adversarial Networks \citep{goodfellow_generative_2014} involve the use of a discriminative model to help a generative model match its output to an existing data distribution via an adversarial minimax game. Such models have achieved good results on various generative tasks such as image synthesis \citep{zhang_self-attention_2018} and text generation \citep{yu_seqgan:_2016}. Recently, adversarial schemes have begun to be adapted for non-strictly-generative tasks such as fake review detection \citep{aghakhani_detecting_2018}, improving the robustness of predictive models to adversarial attacks \citep{zhao_generating_2017} and image retrieval \citep{song_binary_2017}.

Ideas similar to the adversarial scheme used in this paper have come not from interpretability but rather from weakly-supervised object localization. \citet{wei_object_2017} uses a similar scheme to accomplish more complete detection of object shapes in images by iteratively erasing the regions that a predictive model lends the most attention, and forcing it to adjust to the occluded image.

\begin{figure*}[t]
\centering
\includegraphics[width=5in]{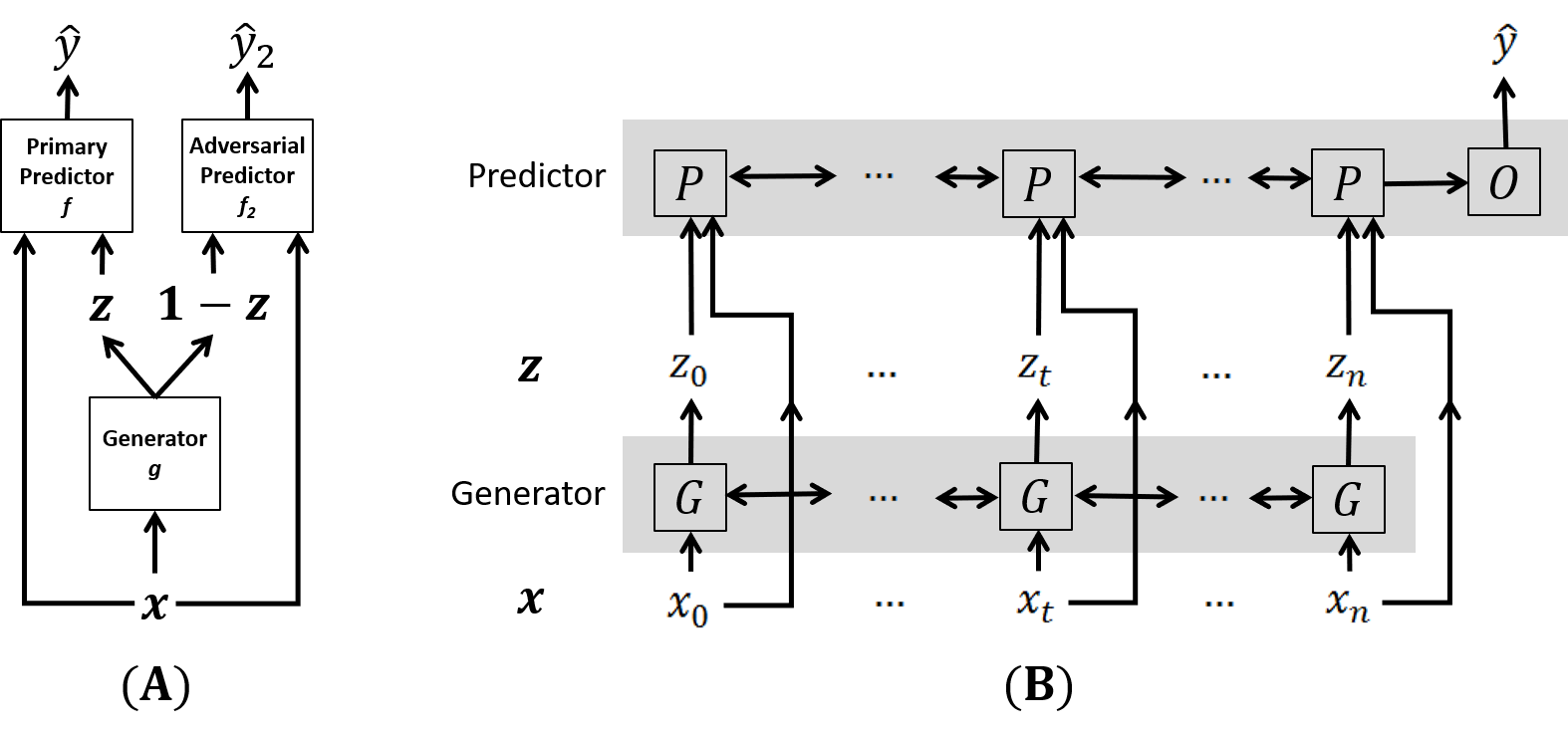}
\caption{(A) Overall architecture. Generator and predictors are RNNs; (B) Detail of interaction between generator and one predictor layer. $G$ and $P$ are recurrent units of any kind. $O$ is a sigmoid output layer.}
\label{fig:combined_architecture}
\end{figure*}

\section{Model}
The goal of our architecture is to highlight personal attacks in text when such are present, and to highlight little or nothing when there are none, while also performing accurate overall prediction. 

These requirements prompt two important edge cases: first, there may be no particular predictive signal in the comment text (i.e. no personal attacks); in a more typical explanatory setting there is always assumed to be some explanation for a decision. Second, there may be redundant signal (i.e. multiple personal attacks), more than is strictly required for accurate prediction, and we assume that it is desirable to identify all of it. We  address both of these cases with modifications to the original model architecture.

The model (Figure \ref{fig:combined_architecture}A) is a hard attention architecture which uses one RNN to extract an attention mask of either 0 or 1 for each token, and a different RNN to make a prediction from the attention-masked text (detailed in Figure \ref{fig:combined_architecture}B). Following \citep{lei_rationalizing_2016}, we refer to the mask-producing layer \textit{g} as the \textit{generator}, but for clarity we call the predictive layer \textit{f}  the \textit{predictor} rather than the encoder. Again following previous work, we refer to the output \textit{z} of the generator as the \textit{rationale}, in that it rationalizes the prediction of the predictor. We also refer to the inverse rationale, defined as 1-\textit{z} , as the \textit{antirationale}. 

To this basic two-layer scheme, we add a secondary, adversarial predictor \textit{$f_2$}, which views the text masked by the antirationale rather than the rationale. The secondary predictor's role is to act as an adversarial discriminator--it tries to make accurate predictions on the antirationale, while the generator tries to prevent it from doing so, which ensures that all predictive signal ends up in the rationale.

\subsection{Primary predictor}

The primary predictor \textit{f} is an RNN which views the input text masked by the rationale produced by the generator. Its objective is simply to reduce its own squared loss: 
\begin{equation}
cost_f(z,x,y) = \big[f(x,z)-y\big]^2
\end{equation}

\begin{figure*}[t]
\centering
\includegraphics[width=5in]{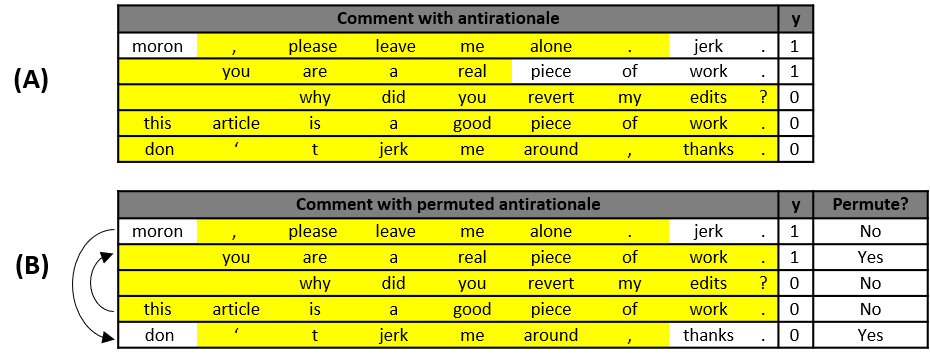}
\caption{(A) Fabricated sample batch masked by antirationales. Note the correlation between mask and target; \\ (B) The batch with some antirationales switched with those of other items. The correlation no longer holds.}
\label{fig:combined_fake_examples}
\end{figure*}

\subsubsection{Default behavior via predictor bias term manipulation}

The default behavior of the model is the prediction the predictor makes if the input is entirely masked by the rationale: $f(x,0)$. When working with a recurrent unit that has no internal bias term, this behavior is entirely determined by the bias term of the final sigmoid output layer, $\sigma(wx+b)$, which with typical random initialization  of \textit{b} results in a default predicted value of roughly 0.5.  

However, this 0.5 default value is not always optimal or semantically appropriate to the predictive task. In the personal attack detection task, if no attacks can be detected, the ``natural'' default target value for a text should be close to 0. We show in the experiments that manually setting the output layer bias term \textit{b} to $logit(0.05)=-2.94$, so that the default predicted value is 0.05, improves model performance.

\subsection{Secondary adversarial predictor}
The secondary adversarial predictor is an RNN which views the input text masked by the antirationale, defined as 1 minus the rationale \textit{z}. Its purpose is to encourage high-recall explanations by trying to make accurate predictions from the antirationale, while the generator tries to prevent it from doing so. 

However, if the adversarial predictor's objective function were simply $\big[f_2(x,1-z)-y\big]^2$, it would be able to gain an unfair advantage from the presence of masking in the antirationale. Seeing evidence of "blanked-out" tokens would tell it that personal attacks were present in that comment, giving it strong hint that the target value is close to 1.0 and vice-versa (see figure \ref{fig:combined_fake_examples}A). 

To take away this advantage, the input to the adversarial predictor has to be permuted such that the mask itself is no longer correlated with the target value, while still allowing it to scan the antirationale for residual predictive signal.

Our solution is to replace the masks of half the items in a training batch with the masks of other items in the batch. We order the batch by target value. If item $x_i$ is selected for replacement, it gets the mask of item $x_{N-i}$ where N is the size of the batch. We call this permutation function $c$:
\begin{equation*}
\begin{split}
&c(z_i) = c(g(x_i)) = 
\begin{cases}
g(x_i) & \text{if } k_i = 1\\
g(x_{N-i}) & \text{if } k_i = 0\\
\end{cases} \\
&x_i \in \{x_0,...,x_N\} \qquad k_i \sim Bernoulli(0.5)
\end{split}
\end{equation*}

This ensures that low-target-value items get masks associated with high target values and vice-versa, to maximize the dissociation between masks and target values. Figure \ref{fig:combined_fake_examples}B demonstrates an example of such permutation. This may slow down the learning, since the adversarial predictor will sometimes have access to somewhat different features of the input than it will have on the test data, but it should not lead to incorrect learning, since the training data always has the correct label, regardless of the mask.

With $c(1-z)$ as the permuted antirationale resulting from applying this randomization process. The objective for the secondary, adversarial predictor is its predictive accuracy on this permuted antirationale:
\begin{equation}
cost_{f_2}(z,x,y) = \big[f_2(x,c(1-z))-y\big]^2
\end{equation}

\subsection{Generator}

Given that the two predictors are trying to minimize error on the rationale and (permuted) antirationale respectively, the objective function for the generator is as follows:
\begin{subequations}
\begin{align}
cost_g&(z,x,y)= \tag{3}\\
  &\big[f(x,z)-y\big]^2  \tag{3.1}\\
    +&\lambda_1 ||z|| \tag{3.2}\\
    +&\lambda_1\lambda_2 \sum_{t}^{} |z_t - z_{t-1}|  \tag{3.3}\\
    +&\lambda_3 \big[f_2(x, 1-z)-f_2(x,0)\big]^2 \tag{3.4}
\end{align}
\end{subequations}

Terms 3.1-3.3 are present in the model of Lei et al. Term 3.1 encourages the generator to allow the primary predictor to make accurate predictions, prevents it from obscuring any tokens that would prevent the predictor from doing so. Term 3.2 encourages the generator to produce minimal rationales; obscuring as many tokens as possible. Term 3.3 encourages rationale coherence by punishing the number of transitions in the rationale; it encourages few contiguous phrases rather than many fragments in the rationale. 

In theory, these three terms ensure high precision, selecting the minimal (term 3.2) rationale with sufficient signal for accurate prediction (term 3.1), subject to a coherence constraint (term 3.3). 

Term 3.4, which is new, ensures recall by encouraging the adversarial predictor's prediction on the antirationale to be similar to the prediction it would make with no information at all (aka the default value). That is, \textbf{the antirationale should contain no predictive signal}. Any personal attacks left out of the rationale would appear in the antirationale, letting the adversarial predictor make a more accurate prediction, which would be penalized by term 3.4.

\subsection{Extractive Adversarial Network}
In the GAN framework \citep{goodfellow_generative_2014}, a discriminator attempts to accurately classify synthetic examples which a generator is striving to match to the distribution of the true data. In our framework, the adversarial predictor attempts to accurately classify censored examples which the generator is striving to strip of all predictive signal. The discriminator in the GAN framework is trained half on real data, and half on fakes; our adversarial predictor is trained half on correctly-masked items and half on items with permuted masks. 
Where our framework differs from GAN is instead of generating adversarial examples which are compared to true examples, our architecture extracts a modified example out of an existing example, and so can therefore be described as an Extractive Adversarial Network (EAN).

\subsection{Implementation details}
For comparability with the original algorithm, we use the same recurrent unit (RCNN) and REINFORCE-style policy gradient optimization process \citep{williams_simple_1992} as \citet{lei_rationalizing_2016} to force the generator outputs to be a discrete 0 or 1. In this framework, the continuous output of the generator on each token is treated as a probability from which the mask is then sampled to produce a discrete value for each token. The gradient across this discontinuity is approximated as:
\begin{equation*}
\begin{aligned}
  &\frac{\partial\textrm{E}_{z \sim g(x)}[cost_g(z,x,y)]}{\partial\theta_{g}} \\
    &=\textrm{E}_{z \sim g(x)}\Bigg[cost_g(z,x,y)\frac{\partial\log p(z|x)}{\partial\theta_{g}}\Bigg]
\end{aligned}
\end{equation*}

In theory, one would sample \textit{z} several times from the generator \textit{g} to produce a good estimate of the gradient. In practice, we find that a single sample per epoch is sufficient. The predictors $f$ and $f_2$ are trained as normal, as the error gradient with respect to their parameters is smooth. 

We employ a particular hard attention model, but the idea of an adversarial critic is not limited to either hard attention or any particular recurrent unit. In a soft attention setting, our adversarial scheme will actually encourage "harder" attention by encouraging any non-zero attention weight to go to 1.0 (or else the inverse of that weight will leave predictive signal in the anti-explanation). 

The attention weights produced by the generator are applied to the predictor at the output rather than the input level. When the recurrent unit $P$ of the predictor operates on a token $x_t$ modified by attention weight $z_t$, it ingests $x_t$ normally, but depending on $z_t$ it either produces its own output or forwards that of the previous token: $$P(x_t,z_t) = z_t P_{base}(x_t) \cdot (1-z_t)P_{base}(x_{t-1})$$

We investigate a similar range of sparsity hyperparameter values as the original model \footnote{$\lambda_1$=[0.0003, 0.0006, 0.0009, 0.0012, 0.0015, 0.0018, 0.0021], $\lambda_2$=[0, 1, 2]}. The weight on the inverse term only matters relative to the model sparsity, as that term cooperates rather than competing with the predictive accuracy term (because it almost never hurts accuracy to add more to the rationale). Therefore we set $\lambda_3$ to 1.0 when we want to include the inverse term. 

We use Word2Vec \citep{mikolov_distributed_2013} to create input token word vectors and Adam \citep{kingma_adam:_2014} for optimization.

\begin{table*}[!ht]
\small
\centering

\begin{tabular}{@{}lccccccccc@{}}
\toprule
\multicolumn{1}{c}{\multirow{3}{*}{\textbf{Model}}} & \multicolumn{6}{c}{\textbf{Rationale}}                                                  & \multicolumn{3}{c}{\multirow{2}{*}{\textbf{Prediction}}} \\ \cmidrule(lr){2-7}
\multicolumn{1}{c}{}                                & \multicolumn{3}{c}{\textbf{Tokenwise}}     & \multicolumn{3}{c}{\textbf{Phrasewise}}    & \multicolumn{3}{c}{}                                     \\ \cmidrule(l){2-10} 
\multicolumn{1}{c}{}                                & \textbf{F1} & \textbf{Pr.} & \textbf{Rec.} & \textbf{F1} & \textbf{Pr.} & \textbf{Rec.} & \textbf{MSE}      & \textbf{Acc.}      & \textbf{F1}     \\ \midrule
Sigmoid predictor                                   & -           & -            & -             & -           & -            & -             & 0.029             & 0.94               & 0.74            \\
RNN predictor                                       & -           & -            & -             & -           & -            & -             & 0.018             & 0.95               & 0.78            \\ \midrule
Mean human performance                              & 0.55        & 0.62         & 0.57          & 0.72        & 0.78         & 0.69          & -                 & -                  & -               \\ \midrule
Sigmoid predictor + feature importance              & 0.20        & \textbf{0.62}         & 0.12          & \textbf{0.64}        & \textbf{0.59}         & 0.70          & 0.029             & 0.94               & 0.74            \\
RNN predictor + sigmoid generator                   & 0.29        & 0.22         & 0.45          & 0.31        & 0.19          & 0.92          & 0.038             & 0.91               & 0.70            \\
RNN predictor + LIME                                & 0.33        & 0.29         & 0.39          & 0.4         & 0.25         & \textbf{0.96}          & 0.018             & 0.95               & 0.78            \\ \midrule
Lei2016                                             & 0.44        & 0.38         & 0.52          & 0.51        & 0.38         & 0.83          & 0.021             & 0.95               & 0.77            \\
Lei2016 + bias                                      & 0.49        & 0.48         & 0.49          & 0.60        & 0.46         & 0.86          & 0.02              & 0.95               & 0.77            \\
Lei2016 + bias + inverse (EAN)                      & \textbf{0.53}        & 0.48         & \textbf{0.58}          & 0.61        & 0.47         & 0.87          & 0.021             & 0.95               & 0.77            \\ \bottomrule
\end{tabular}
\caption{Rationale performance relative to human annotations. Prediction accuracy is based on a binary threshold of 0.5. Performance of both Lei2016 model variants is significantly different from the baseline model (McNemar's test, $p < 0.05$)}
\label{table:rationale_performance_results}

\end{table*}

\section{Data}
To train our model of personal attacks, we use the dataset introduced by \citep{wulczyn_ex_2017}, which consists of roughly 100,000 Wikipedia revision comments labeled via crowsourcing for aggression, toxicity and the presence of personal attacks. This dataset includes its own training, development and test set split, which we also use. 

To this dataset we add a small validation set of personal attack rationales. 40 undergraduate students used Brat \citep{stenetorp_brat:_2012} to highlight sections of comments that they considered to constitute personal attacks. Comments were sampled in a stratified manner from the development and test sets of the Wulczyn et al. dataset, and each student annotated roughly 150 comments, with each comment viewed by roughly 4 annotators. To calculate gold-standard rationales, we take the majority vote among annotators for each token in each comment. 1089 distinct comments were annotated, split between a development and test set of 549 and 540 examples respectively. 

The Krippendorff's alpha on our validation set is 0.53 at the whole-comment level. This value is comparable with that of \citet{wulczyn_ex_2017} (0.45). Agreement at the token level is a lower 0.41, because this includes tokens which are a matter of preference among annotators, such as articles and adverbs, as well as content tokens.

\section{Experiments}
We show that both modifications to the original algorithm, bias term manipulation and inverse term, increase the tokenwise F1 of the predicted rationales relative to our human-annotated test set.  
All hyperparameters were tuned to maximize tokenwise F1 on the development set. \footnote{$\lambda_1$=0.0006 for variants without inverse term, $\lambda_1$=0.0015 for variant with inverse term, $\lambda_2$=2 (Tuned for maximum F1 on original model, then held constant for comparability)}

\subsection{Baselines}
We generate six baselines for comparison with our variant of the \citep{lei_rationalizing_2016} architecture. These include the following:

\noindent\textbf{Sigmoid predictor (logistic regression):} Bag-of-words representation with a sigmoid output layer.

\noindent\textbf{RNN predictor:} The same sequence model used for the predictor, but with no generator layer.

\noindent\textbf{Mean human performance:} The mean tokenwise performance of human annotators measured against the majority vote for the comments they annotated (with their vote left out).

\noindent\textbf{Sigmoid predictor + feature importance:} Bag-of-words representation with sigmoid output layer, with post-hoc feature importance based on model coefficients. Cutoff threshold for features tuned to maximize  rationale F1 on development set.

\noindent\textbf{RNN predictor + sigmoid generator:} Rationale mask generated by sigmoid layer applied independently to each input token. Prediction layer is same as predictor.

\noindent\textbf{RNN predictor + LIME:} Rationale mask generated by applying LIME \citep{ribeiro_why_2016} post-hoc to RNN layer predictions. Masking threshold tuned to maximize rationale F1.  

\subsection{Rationale performance}
In the main experiment, we evaluate model rationales relative to rationales created by human annotators. In our validation dataset, human annotators typically chose to annotate personal attacks at the phrase level; hence in the sentence ``Get a job, you hippie s***bag'', the majority-vote rationale consists of the entire sentence, where it could arguably consist of the last two or even the last word. Therefore, in addition to tokenwise precision, recall and positive F1, we also report a relaxed ``phrasewise'' version of these metrics where any time we capture part of a contiguous rationale chunk, that is considered a true positive.  

We report results for the original model (i.e. terms 3.1-3.3 in the objective function), the original model with its bias term set for a default value of 0.05, and the bias-modified model with the additional inverse term (term 3.4). For every model variant, we optimized hyperparameters for tokenwise F1 on the development set. We also report results for the baselines described above.

Table 1 displays the results. The difference in performance between the three baselines that don't use a RNN generator and the three model variants that do demonstrates the importance of context in recognizing personal attacks within text. The relative performance of the three variants of the Lei et al. model show that both modifications, setting the bias term and the addition of the adversarial predictor, lead to marginal improvements in tokenwise F1. The best-performing model approaches average human performance on this metric. 

The phrasewise metric is relaxed. It allows a contiguous personal attack sequence to be considered captured if even a single token from the sequence is captured. The results on this metric show that in an absolute sense, 87\% of personal attacks are at least partially captured by the algorithm. The simplest baseline, which produces rationales by thresholding the coefficients of a logistic regression model, does deceptively well on this metric by only identifying attacking words like "jerk" and "a**hole", but its poor tokenwise performance shows that it doesn't mimic human highlighting very well.

\subsection{Original model tokenwise recall}

A perplexing result of the rationale performance comparison is how good the tokenwise recall of the model is \textit{without} the inverse term. Without it, the model is encouraged to find the minimal rationale which offers good predictive performance. Comments with more than one personal attack (e.g. Figure \ref{fig:high_attacking_example}) constitute 29\% of those with at least one attack and 13\% of all comments in our validation set. For comments like these, the model should in theory only identify one such attack. However, it tends to find more information than needed, leading to a higher-than-expected recall of .52 in the best overall version of this variant.

To explain this behavior, we run a leave-one-out experiment on the original+bias and original+bias+inverse model variants. For each distinct contiguous rationale chunk predicted by each model (when it generates multi-piece rationales), we try removing this chunk from the predicted rationale, running the prediction layer on the reduced rationale, and seeing whether the result lowers the value of the overall objective function. 

For the original+bias model variant, we find that performing this reduction improves the value of the objective function 65\% of the time. However, the combined average impact of these reductions on the objective function is to worsen it. What this means is while 65\% of distinct phrases discovered by the generator are unnecessary for accurate prediction, the 35\% of them that are necessary lead to a major decrease in predictive accuracy. 

That is, the generator ``hedges its bets'' with respect to predictive accuracy by including more information in the rationales than it has to, and experiences a better global optimum as a result. This behavior is less prominent with the inclusion of the inverse term, where the percentage of unnecessary rationale phrases falls to 47\%.

\begin{figure}[t]
\includegraphics[width=3in]{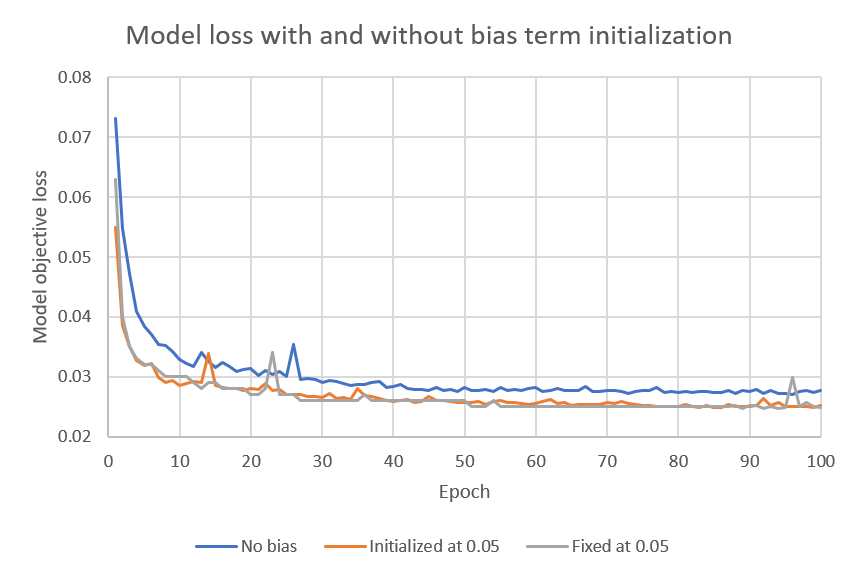}
\caption{Evolution of model loss over time with and without bias term manipulation}
\label{fig:generator_loss_bias}
\end{figure}

\begin{figure}[t]
\includegraphics[width=3in]{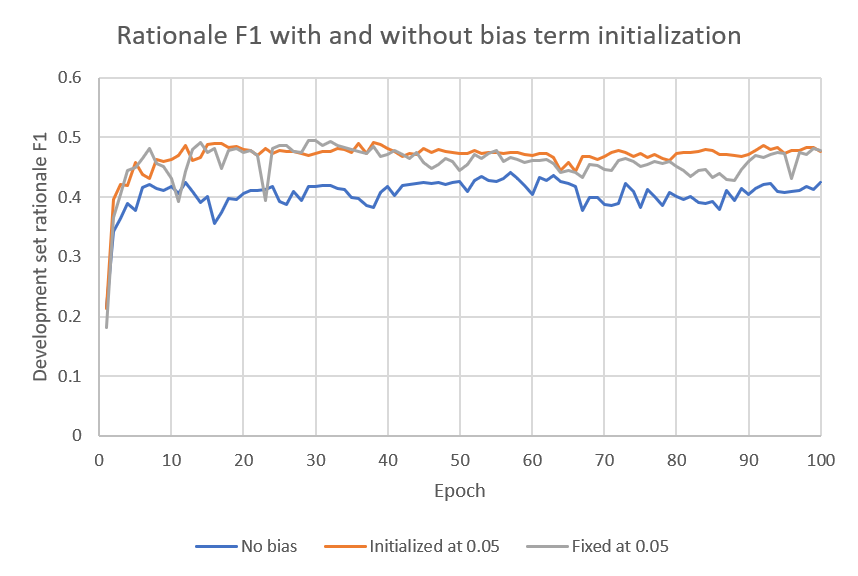}
\caption{Evolution of development set rationale F1 score over time with and without bias term manipulation}
\label{fig:rationale_f1_bias}
\end{figure}

\begin{figure*}[!ht]
\centering
\includegraphics[width=6.5in]{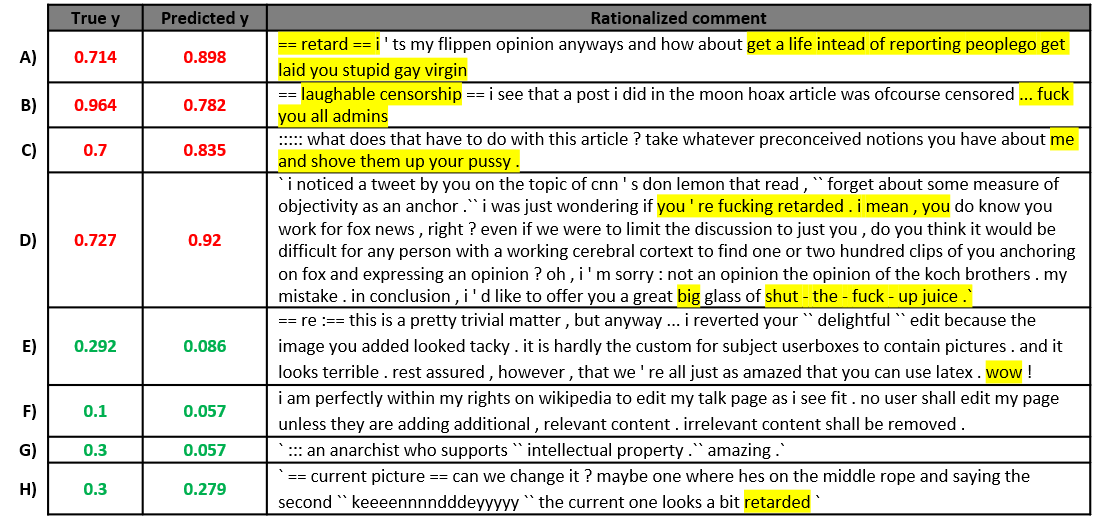}
\caption{Further examples of labeled and rationalized comments. Items E) and G) show that the algorithm struggles with sarcasm.}
\label{fig:more_examples}
\end{figure*}
\subsection{Impact of bias term manipulation}

In theory, the model should learn a good bias term for the predictor layer, and therefore the idea of explicitly initializing or fixing the bias term to match the semantics of the task should not impact model performance or represent much of a contribution.

In practice however, as figures \ref{fig:generator_loss_bias} and \ref{fig:rationale_f1_bias} demonstrate, the initialization of the bias term has a big impact on even the long-term learning behavior of the model. Using the best hyperparameters for the original no-bias, no-inverse-term model, figure \ref{fig:generator_loss_bias} shows that either initializing or permanently fixing the predictor bias for a default output value of 0.05 leads to improved model loss with respect to its own objective function. Figure \ref{fig:rationale_f1_bias} shows a similar pattern for tokenwise F1 score.

\section{Discussion and future work}

One interpretation of the impact of the bias term on model behavior is that an explanation of ``why'' is really an explanation of ``why not''--that is, an explanation is information that distinguishes an item from some alternative hypothesis, and explicitly choosing what this alternative is can improve explanation performance (particularly precision).

Manually setting the model to produce some reasonable default value for an empty rationale makes sense in our setting, but not in domains where there is no default value, such as the beer review dataset of \citep{lei_rationalizing_2016}. A more general approach would be to base explanations on confidence rather than accuracy, where the default value would simply be the mean and variance of the training data, and explanations would consist of tokens that tighten the bounds on the output.
  
A surprising finding is that the original algorithm often ends up defying its own objective and finds more complete rationales than needed. The leave-one-out experiment described above suggests that the reason for this behavior is that it is how the generator deals with predictive uncertainty, and that it achieves a better global optimum by producing locally suboptimal rationales.

While this ``bug'' proves useful in our case, it may not generalize. In our setting the adversarial predictor gives a modest improvement in recall; it will produce a larger improvement in settings where the unaltered algorithm is more successful at producing the minimal explanations described by its objective function. \citet{li_understanding_2016} finds that a memory network predictor requires less occlusion than an LSTM to flip its predictions, indicating that choice of model can effect completeness of explanations. 

In theory, interpretable models can aid human moderaters by pointing them directly at the potentially objectionable content in a comment and giving them a starting point for making their own holistic decision about the comment. However, there are potential pitfalls. Adding explanations as a model output gives the model another way to be wrong--one which humans may be even less able to troubleshoot than simple misclassification. Relatedly, explanations may inspire overconfidence in model predictions. Extensive user testing would clearly be needed before any deployment.

One final concern is the question of whether human-like explanations are really optimal explanations. Are high-recall explanations that mimic human highlighting tendencies really optimal for the types of moderating/self-moderating tasks involved in the domain of personal attacks in online social media? Again, this question can only be answered with human subject experimentation, which we plan to approach in future work.

\section{Conclusion}

The main contribution of this paper is to frame explanation as an adversarial problem, thereby addressing explanation recall for the first time that we are aware of. We do so by introducing an adversarial framework (an ``extractive adversarial network'') for ensuring that redundant predictive signal is not omitted from a model's explanations. We also show that choosing a null hypothesis for the model by setting the model bias term improves explanation precision. 

Secondarily, we make a domain-specific contribution by applying interpretable machine learning for the first time to the problem of identifying personal attacks in social media comments, with the hope of developing more transparent semi-automated moderation systems. We show that we approach human performance on a dataset we develop for this purpose.

\section*{Acknowledgement}
This material is based upon work supported by the National Science Foundation under grant numbers 1717688, 1633370 and 1620319. We also thank David Jurgens and Yue Wang for their helpful comments and suggestions. 

\bibliography{emnlp}

\begin{thebibliography}{43}
\expandafter\ifx\csname natexlab\endcsname\relax\def\natexlab#1{#1}\fi

\bibitem[{Abbott et~al.(2016)Abbott, Ecker, Anand, and
  Walker}]{abbott_internet_2016}
Robert Abbott, Brian Ecker, Pranav Anand, and Marilyn Walker. 2016.
\newblock Internet {Argument} {Corpus} 2.0: {An} {SQL} {Schema} for {Dialogic}
  {Social} {Media} and the {Corpora} to {Go} {With} {It}.
\newblock In \emph{Language {Resources} and {Evaluation} {Conference} ({LREC}
  2016)}.

\bibitem[{Adams and Dixon(2017)}]{adams_better_2017}
CJ~Adams and Lucas Dixon. 2017.
\newblock Better discussions with imperfect models.
\newblock \emph{The False Positive - Medium}.
\newblock [Blog post].

\bibitem[{Aghakhani et~al.(2018)Aghakhani, Machiry, Nilizadeh, Kruegel, and
  Vigna}]{aghakhani_detecting_2018}
Hojjat Aghakhani, Aravind Machiry, Shirin Nilizadeh, Christopher Kruegel, and
  Giovanni Vigna. 2018.
\newblock Detecting {Deceptive} {Reviews} using {Generative} {Adversarial}
  {Networks}.
\newblock \emph{arXiv:1805.10364 [cs]}.
\newblock ArXiv: 1805.10364.

\bibitem[{Ancona et~al.(2017)Ancona, Ceolini, Öztireli, and
  Gross}]{ancona_towards_2017}
Marco Ancona, Enea Ceolini, Cengiz Öztireli, and Markus Gross. 2017.
\newblock Towards better understanding of gradient-based attribution methods
  for {Deep} {Neural} {Networks}.
\newblock \emph{arXiv:1711.06104 [cs, stat]}.
\newblock ArXiv: 1711.06104.

\bibitem[{Anderson et~al.(2014)Anderson, Brossard, Scheufele, Xenos, and
  Ladwig}]{anderson_nasty_2014}
Ashley~A. Anderson, Dominique Brossard, Dietram~A. Scheufele, Michael~A. Xenos,
  and Peter Ladwig. 2014.
\newblock The “{Nasty} {Effect}:” {Online} {Incivility} and {Risk}
  {Perceptions} of {Emerging} {Technologies}: {Crude} comments and concern.
\newblock \emph{Journal of Computer-Mediated Communication}, 19(3):373--387.

\bibitem[{Anderson et~al.(2016)Anderson, Yeo, Brossard, Scheufele, and
  Xenos}]{anderson_toxic_2016}
Ashley~A. Anderson, Sara~K. Yeo, Dominique Brossard, Dietram~A. Scheufele, and
  Michael~A. Xenos. 2016.
\newblock Toxic {Talk}: {How} {Online} {Incivility} {Can} {Undermine}
  {Perceptions} of {Media}.
\newblock \emph{International Journal of Public Opinion Research}.

\bibitem[{Arras et~al.(2017)Arras, Montavon, Müller, and
  Samek}]{arras_explaining_2017}
Leila Arras, Grégoire Montavon, Klaus-Robert Müller, and Wojciech Samek.
  2017.
\newblock Explaining {Recurrent} {Neural} {Network} {Predictions} in
  {Sentiment} {Analysis}.
\newblock In \emph{Proceedings of the 8th {Workshop} on {Computational}
  {Approaches} to {Subjectivity}, {Sentiment} and {Social} {Media} {Analysis}},
  pages 159--168, Copenhagen, Denmark. Association for Computational
  Linguistics.

\bibitem[{{Association for Computational
  Linguistics}(2017)}]{association_for_computational_linguistics_proceedings_2017}
{Association for Computational Linguistics}. 2017.
\newblock \emph{Proceedings of the 1st {Workshop} on {Abusive} {Language}
  {Online}}.
\newblock ACL, Vancouver.

\bibitem[{Binns et~al.(2017)Binns, Veale, Van~Kleek, and
  Shadbolt}]{binns_like_2017}
Reuben Binns, Michael Veale, Max Van~Kleek, and Nigel Shadbolt. 2017.
\newblock Like {Trainer}, {Like} {Bot}? {Inheritance} of {Bias} in
  {Algorithmic} {Content} {Moderation}.
\newblock In \emph{Social {Informatics}}, volume 10540, pages 405--415, Cham.
  Springer.

\bibitem[{Blackwell et~al.(2018)Blackwell, Dimond, Schoenebeck, and
  Lampe}]{blackwell_classification_2018}
Lindsay Blackwell, Jill Dimond, Sarita Schoenebeck, and Cliff Lampe. 2018.
\newblock Classification and {Its} {Consequences} for {Online} {Harassment}:
  {Design} {Insights} from {HeartMob}.
\newblock In \emph{Proceedings of the {ACM} {Conference} on {Computer}
  {Supported} {Cooperative} {Work} ({CSCW} 2018)}, pages 1--19.

\bibitem[{Cheng et~al.(2015)Cheng, Danescu-Niculescu-Mizil, and
  Leskovec}]{cheng_antisocial_2015}
Justin Cheng, Cristian Danescu-Niculescu-Mizil, and Jure Leskovec. 2015.
\newblock Antisocial {Behavior} in {Online} {Discussion} {Communities}.
\newblock In \emph{Proceedings of the {International} {Conference} on {Web} and
  {Social} {Media}}, page~10.

\bibitem[{Doshi-Velez and Kim(2017)}]{doshi-velez_towards_2017-1}
Finale Doshi-Velez and Been Kim. 2017.
\newblock Towards {A} {Rigorous} {Science} of {Interpretable} {Machine}
  {Learning}.
\newblock \emph{arXiv:1702.08608 [cs, stat]}.
\newblock ArXiv: 1702.08608.

\bibitem[{Ehsan et~al.(2017)Ehsan, Harrison, Chan, and
  Riedl}]{ehsan_rationalization:_2017}
Upol Ehsan, Brent Harrison, Larry Chan, and Mark~O. Riedl. 2017.
\newblock Rationalization: {A} {Neural} {Machine} {Translation} {Approach} to
  {Generating} {Natural} {Language} {Explanations}.
\newblock \emph{arXiv:1702.07826 [cs]}.
\newblock ArXiv: 1702.07826.

\bibitem[{Gilpin et~al.(2018)Gilpin, Bau, Yuan, Bajwa, Specter, and
  Kagal}]{gilpin_explaining_2018}
Leilani~H. Gilpin, David Bau, Ben~Z. Yuan, Ayesha Bajwa, Michael Specter, and
  Lalana Kagal. 2018.
\newblock Explaining {Explanations}: {An} {Approach} to {Evaluating}
  {Interpretability} of {Machine} {Learning}.
\newblock \emph{arXiv:1806.00069 [cs, stat]}.
\newblock ArXiv: 1806.00069.

\bibitem[{Golbeck et~al.(2017)Golbeck, Gnanasekaran, Gunasekaran, Hoffman,
  Hottle, Jienjitlert, Khare, Lau, Martindale, Naik, Nixon, Ashktorab,
  Ramachandran, Rogers, Rogers, Sarin, Shahane, Thanki, Vengataraman, Wan, Wu,
  Banjo, Berlinger, Bhagwan, Buntain, Cheakalos, Geller, and
  Gergory}]{golbeck_large_2017}
Jennifer Golbeck, Rajesh~Kumar Gnanasekaran, Raja~Rajan Gunasekaran, Kelly~M.
  Hoffman, Jenny Hottle, Vichita Jienjitlert, Shivika Khare, Ryan Lau,
  Marianna~J. Martindale, Shalmali Naik, Heather~L. Nixon, Zahra Ashktorab,
  Piyush Ramachandran, Kristine~M. Rogers, Lisa Rogers, Meghna~Sardana Sarin,
  Gaurav Shahane, Jayanee Thanki, Priyanka Vengataraman, Zijian Wan,
  Derek~Michael Wu, Rashad~O. Banjo, Alexandra Berlinger, Siddharth Bhagwan,
  Cody Buntain, Paul Cheakalos, Alicia~A. Geller, and Quint Gergory. 2017.
\newblock A {Large} {Labeled} {Corpus} for {Online} {Harassment} {Research}.
\newblock In \emph{Proceedings of the 2017 {ACM} on {Web} {Science}
  {Conference}}, pages 229--233. ACM Press.

\bibitem[{Goodfellow et~al.(2014)Goodfellow, Pouget-Abadie, Mirza, Xu,
  Warde-Farley, Ozair, Courville, and Bengio}]{goodfellow_generative_2014}
Ian Goodfellow, Jean Pouget-Abadie, Mehdi Mirza, Bing Xu, David Warde-Farley,
  Sherjil Ozair, Aaron Courville, and Yoshua Bengio. 2014.
\newblock Generative {Adversarial} {Nets}.
\newblock In \emph{Advances in {Neural} {Information} {Processing} {Systems}},
  pages 2672--2680.

\bibitem[{Guidotti et~al.(2018)Guidotti, Monreale, Turini, and
  Pedreschi}]{guidotti_survey_2018}
Riccardo Guidotti, Anna Monreale, Franco Turini, and Dino Pedreschi. 2018.
\newblock A {Survey} {Of} {Methods} {For} {Explaining} {Black} {Box} {Models}.
\newblock \emph{arXiv preprint arXiv:1802.01933}.

\bibitem[{Hosseini et~al.(2017)Hosseini, Kannan, Zhang, and
  Poovendran}]{hosseini_deceiving_2017}
Hossein Hosseini, Sreeram Kannan, Baosen Zhang, and Radha Poovendran. 2017.
\newblock Deceiving {Google}’s {Perspective} {API} {Built} for {Detecting}
  {Toxic} {Comments}.
\newblock \emph{arXiv preprint arXiv:1702.08138}.

\bibitem[{Kennedy et~al.(2017)Kennedy, McCollough, Dixon, Bastidas, Ryan, Loo,
  and Sahay}]{kennedy_technology_2017}
George Kennedy, Andrew McCollough, Edward Dixon, Alexei Bastidas, John Ryan,
  Chris Loo, and Saurav Sahay. 2017.
\newblock Technology {Solutions} to {Combat} {Online} {Harassment}.
\newblock In \emph{Proceedings of the {First} {Workshop} on {Abusive}
  {Language} {Online}}, pages 73--77. Association for Computational
  Linguistics.

\bibitem[{Kim et~al.(2017)Kim, Wattenberg, Gilmer, Cai, Wexler, Viegas, and
  Sayres}]{kim_interpretability_2017}
Been Kim, Martin Wattenberg, Justin Gilmer, Carrie Cai, James Wexler, Fernanda
  Viegas, and Rory Sayres. 2017.
\newblock Interpretability {Beyond} {Feature} {Attribution}: {Quantitative}
  {Testing} with {Concept} {Activation} {Vectors} ({TCAV}).
\newblock \emph{arXiv:1711.11279 [stat]}.
\newblock ArXiv: 1711.11279.

\bibitem[{Kingma and Ba(2014)}]{kingma_adam:_2014}
Diederik~P. Kingma and Jimmy Ba. 2014.
\newblock Adam: {A} {Method} for {Stochastic} {Optimization}.
\newblock \emph{Proceedings of the 3rd International Conference on Learning
  Representations}.

\bibitem[{Lakkaraju et~al.(2016)Lakkaraju, Bach, and
  Leskovec}]{lakkaraju_interpretable_2016}
Himabindu Lakkaraju, Stephen~H. Bach, and Jure Leskovec. 2016.
\newblock Interpretable {Decision} {Sets}: {A} {Joint} {Framework} for
  {Description} and {Prediction}.
\newblock In \emph{Proceedings of the 22nd {ACM} {SIGKDD} {International}
  {Conference} on {Knowledge} {Discovery} and {Data} {Mining}}, pages
  1675--1684. ACM Press.

\bibitem[{Lei et~al.(2016)Lei, Barzilay, and Jaakkola}]{lei_rationalizing_2016}
Tao Lei, Regina Barzilay, and Tommi Jaakkola. 2016.
\newblock Rationalizing {Neural} {Predictions}.
\newblock In \emph{Proceedings of the 2016 {Conference} on {Empirical}
  {Methods} in {Natural} {Language} {Processing}}, pages 107--117.

\bibitem[{Lenhart et~al.(2016)Lenhart, Ybarra, Zickuhr, and
  Price-Feeney}]{lenhart_online_2016}
Amanda Lenhart, Michele Ybarra, Kathryn Zickuhr, and Myeshia Price-Feeney.
  2016.
\newblock Online {Harassment}, {Digital} {Abuse}, and {Cyberstalking} in
  {America}.
\newblock Technical report, Data \& Society Research Institute.

\bibitem[{Li et~al.(2016)Li, Monroe, and Jurafsky}]{li_understanding_2016}
Jiwei Li, Will Monroe, and Dan Jurafsky. 2016.
\newblock Understanding {Neural} {Networks} through {Representation} {Erasure}.
\newblock \emph{arXiv preprint arXiv:1612.08220}.

\bibitem[{Li et~al.(2017)Li, Liu, Chen, and Rudin}]{li_deep_2017-1}
Oscar Li, Hao Liu, Chaofan Chen, and Cynthia Rudin. 2017.
\newblock Deep {Learning} for {Case}-{Based} {Reasoning} through {Prototypes}:
  {A} {Neural} {Network} that {Explains} {Its} {Predictions}.
\newblock \emph{arXiv:1710.04806 [cs, stat]}.
\newblock ArXiv: 1710.04806.

\bibitem[{Mikolov et~al.(2013)Mikolov, Sutskever, Chen, Corrado, and
  Dean}]{mikolov_distributed_2013}
Tomas Mikolov, Ilya Sutskever, Kai Chen, Greg~S. Corrado, and Jeff Dean. 2013.
\newblock Distributed representations of words and phrases and their
  compositionality.
\newblock In \emph{Advances in neural information processing systems}, pages
  3111--3119.

\bibitem[{Napoles et~al.(2017)Napoles, Tetreault, Pappu, Rosato, and
  Provenzale}]{napoles_finding_2017}
Courtney Napoles, Joel Tetreault, Aasish Pappu, Enrica Rosato, and Brian
  Provenzale. 2017.
\newblock Finding {Good} {Conversations} {Online}: {The} {Yahoo} {News}
  {Annotated} {Comments} {Corpus}.
\newblock In \emph{Proceedings of the 11th {Linguistic} {Annotation}
  {Workshop}}.

\bibitem[{Nguyen(2018)}]{nguyen_comparing_2018}
Dong Nguyen. 2018.
\newblock Comparing {Automatic} and {Human} {Evaluation} of {Local}
  {Explanations} for {Text} {Classification}.
\newblock In \emph{Proceedings of the 2018 {Conference} of the {North}
  {American} {Chapter} of the {Association} for {Computational} {Linguistics}:
  {Human} {Language} {Technologies}}, pages 1069--1078.

\bibitem[{Nobata et~al.(2016)Nobata, Tetreault, Thomas, Mehdad, and
  Chang}]{nobata_abusive_2016}
Chikashi Nobata, Joel Tetreault, Achint Thomas, Yashar Mehdad, and Yi~Chang.
  2016.
\newblock Abusive {Language} {Detection} in {Online} {User} {Content}.
\newblock In \emph{Proceedings of the 25th {International} {Conference} on
  {World} {Wide} {Web}}, pages 145--153.

\bibitem[{Pavlopoulos et~al.(2017)Pavlopoulos, Malakasiotis, and
  Androutsopoulos}]{pavlopoulos_deep_2017}
John Pavlopoulos, Prodromos Malakasiotis, and Ion Androutsopoulos. 2017.
\newblock Deep {Learning} for {User} {Comment} {Moderation}.
\newblock In \emph{Proceedings of the {First} {Workshop} on {Abusive}
  {Language} {Online}}, pages 25--35.

\bibitem[{Pew(2016)}]{pew_political_2016}
Pew. 2016.
\newblock The {Political} {Environment} on {Social} {Media}.
\newblock Technical report, Pew Research Center.

\bibitem[{Ribeiro et~al.(2016)Ribeiro, Singh, and Guestrin}]{ribeiro_why_2016}
Marco~Tulio Ribeiro, Sameer Singh, and Carlos Guestrin. 2016.
\newblock "{Why} {Should} {I} {Trust} {You}?": {Explaining} the {Predictions}
  of {Any} {Classifier}.
\newblock In \emph{Proceedings of the 22nd {ACM} {SIGKDD} {International}
  {Conference} on {Knowledge} {Discovery} and {Data} {Mining}}, pages
  1135--1144. ACM.

\bibitem[{Ribeiro et~al.(2018)Ribeiro, Singh, and
  Guestrin}]{ribeiro_anchors:_2018}
Marco~Tulio Ribeiro, Sameer Singh, and Carlos Guestrin. 2018.
\newblock Anchors: {High} {Precision} {Model}-{Agnostic} {Explanations}.
\newblock In \emph{Proceedings of the {AAAI} {Conference} on {Artificial}
  {Intelligence} ({AAAI})}, page~9.

\bibitem[{Simonyan et~al.(2013)Simonyan, Vedaldi, and
  Zisserman}]{simonyan_deep_2013}
Karen Simonyan, Andrea Vedaldi, and Andrew Zisserman. 2013.
\newblock Deep {Inside} {Convolutional} {Networks}: {Visualising} {Image}
  {Classification} {Models} and {Saliency} {Maps}.
\newblock \emph{arXiv:1312.6034 [cs]}.
\newblock ArXiv: 1312.6034.

\bibitem[{Song(2017)}]{song_binary_2017}
Jingkuan Song. 2017.
\newblock Binary {Generative} {Adversarial} {Networks} for {Image} {Retrieval}.
\newblock \emph{arXiv:1708.04150 [cs]}.
\newblock ArXiv: 1708.04150.

\bibitem[{Stenetorp et~al.(2012)Stenetorp, Pyysalo, Topić, Ohta, Ananiadou,
  and Tsujii}]{stenetorp_brat:_2012}
Pontus Stenetorp, Sampo Pyysalo, Goran Topić, Tomoko Ohta, Sophia Ananiadou,
  and Jun'ichi Tsujii. 2012.
\newblock {BRAT}: a web-based tool for {NLP}-assisted text annotation.
\newblock In \emph{Proceedings of the {Demonstrations} at the 13th {Conference}
  of the {European} {Chapter} of the {Association} for {Computational}
  {Linguistics}}, pages 102--107.

\bibitem[{Wei et~al.(2017)Wei, Feng, Liang, Cheng, Zhao, and
  Yan}]{wei_object_2017}
Yunchao Wei, Jiashi Feng, Xiaodan Liang, Ming-Ming Cheng, Yao Zhao, and
  Shuicheng Yan. 2017.
\newblock Object {Region} {Mining} with {Adversarial} {Erasing}: {A} {Simple}
  {Classification} to {Semantic} {Segmentation} {Approach}.
\newblock \emph{arXiv:1703.08448 [cs]}.
\newblock ArXiv: 1703.08448.

\bibitem[{Williams(1992)}]{williams_simple_1992}
Ronald~J. Williams. 1992.
\newblock Simple statistical gradient-following algorithms for connectionist
  reinforcement learning.
\newblock \emph{Machine learning}, 8(3-4):229--256.

\bibitem[{Wulczyn et~al.(2017)Wulczyn, Thain, and Dixon}]{wulczyn_ex_2017}
Ellery Wulczyn, Nithum Thain, and Lucas Dixon. 2017.
\newblock Ex {Machina}: {Personal} {Attacks} {Seen} at {Scale}.
\newblock In \emph{Proceedings of the 26th {International} {Conference} on
  {World} {Wide} {Web}}, pages 1391--1399.

\bibitem[{Yu et~al.(2016)Yu, Zhang, Wang, and Yu}]{yu_seqgan:_2016}
Lantao Yu, Weinan Zhang, Jun Wang, and Yong Yu. 2016.
\newblock {SeqGAN}: {Sequence} {Generative} {Adversarial} {Nets} with {Policy}
  {Gradient}.
\newblock \emph{AAAI Conference on Artificial Intelligence}.

\bibitem[{Zhang et~al.(2018)Zhang, Goodfellow, Metaxas, and
  Odena}]{zhang_self-attention_2018}
Han Zhang, Ian Goodfellow, Dimitris Metaxas, and Augustus Odena. 2018.
\newblock Self-{Attention} {Generative} {Adversarial} {Networks}.
\newblock \emph{arXiv:1805.08318 [cs, stat]}.
\newblock ArXiv: 1805.08318.

\bibitem[{Zhao et~al.(2017)Zhao, Dua, and Singh}]{zhao_generating_2017}
Zhengli Zhao, Dheeru Dua, and Sameer Singh. 2017.
\newblock Generating {Natural} {Adversarial} {Examples}.
\newblock \emph{arXiv:1710.11342 [cs]}.
\newblock ArXiv: 1710.11342.

\end{thebibliography}

\bibliographystyle{acl_natbib_nourl}

\end{document}